\documentclass[letterpaper, 10 pt, conference]{ieeeconf}  

\IEEEoverridecommandlockouts                              

\title{\LARGE \bf
Design, Contact Modeling, and Collision-inclusive Planning of a Dual-stiffness Aerial RoboT (DART)
}

\author{Yogesh Kumar, Karishma Patnaik and Wenlong Zhang$^*$
\thanks{This material is based upon work supported by the National Science Foundation under Grant No. 2331781.}
\thanks{The authors are with the School of Manufacturing Systems and Networks, Ira A. Fulton Schools of Engineering, Arizona State University, Mesa, AZ, 85212, USA. Email: {\tt\small $\{$ykumar6, kpatnaik, wenlong.zhang$\}$@asu.edu}.}%
\thanks{$^*$Address all correspondence to this author.}%
}
\usepackage{url}
\usepackage{amsmath}
\usepackage{graphicx}
\usepackage{amssymb}
\usepackage{mathtools}
\usepackage{xcolor}
\usepackage{float}
\usepackage{cite}
\usepackage[caption=false]{subfig}
\usepackage{hyperref}
\usepackage{tikz}

\usepackage[bottom=1.1in, top=52pt,right=46pt,left=46pt]{geometry}

\newcommand\copyrighttext{%
	\footnotesize \textcopyright 2025 IEEE. Accepted for publication at IEEE ICRA 2025. Personal use of this material is permitted.
	Permission from IEEE must be obtained for all other uses, in any current or future 
	media, including reprinting/republishing this material for advertising or promotional 
	purposes, creating new collective works, for resale or redistribution to servers or 
	lists, or reuse of any copyrighted component of this work in other works. 
}
\newcommand\copyrightnotice{%
	\begin{tikzpicture}[remember picture,overlay]
		\node[anchor=south,yshift=10pt] at (current page.south) {\fbox{\parbox{\dimexpr\textwidth-\fboxsep-\fboxrule\relax}{\copyrighttext}}};
	\end{tikzpicture}%
}

\begin{document}

\graphicspath{{figures/}}
\maketitle
\copyrightnotice
\thispagestyle{empty}
\pagestyle{empty}


\begin{abstract}
Collision-resilient quadrotors have gained significant attention given their potential for operating in cluttered environments and leveraging impacts to perform agile maneuvers. However, existing designs are typically single-mode: either safeguarded by propeller guards that prevent deformation or deformable but lacking rigidity, which is crucial for stable flight in open environments. This paper introduces DART, a Dual-stiffness Aerial RoboT, that adapts its post-collision response by either engaging a locking mechanism for a rigid mode or disengaging it for a flexible mode, respectively. Comprehensive characterization tests highlight the significant difference in post-collision responses between its rigid and flexible modes, with the rigid mode offering seven times higher stiffness compared to the flexible mode. To understand and harness the collision dynamics, we propose a novel collision response prediction model based on the linear complementarity system theory. We demonstrate the accuracy of predicting collision forces for both the rigid and flexible modes of DART. Experimental results confirm the accuracy of the model and underscore its potential to advance collision-inclusive trajectory planning in aerial robotics.

\end{abstract}

\section{Introduction}\label{sec:intro}
Collision-resilient quadrotors with avian-like compliant bodies that withstand impacts have risen to prominence in robotics, aerodynamics, and biology \cite{PC+22,ML22}. 
Leveraging sophisticated collision-tolerant mechanisms, these drones exhibit remarkable agility and adaptability, making them invaluable assets in various applications requiring physical-interaction with the environment such as navigation in confined environments \cite{PM+20,AA+22}, haptic-based navigation \cite{PS+23,BK+13} and contact-reactive perching \cite{NP+23,ZH24}. 

\begin{figure}[t]
    \centering
    \includegraphics[angle=180, origin=c, trim = 3.0cm 0.5cm 0.5cm 0cm, clip, width=0.45\textwidth]{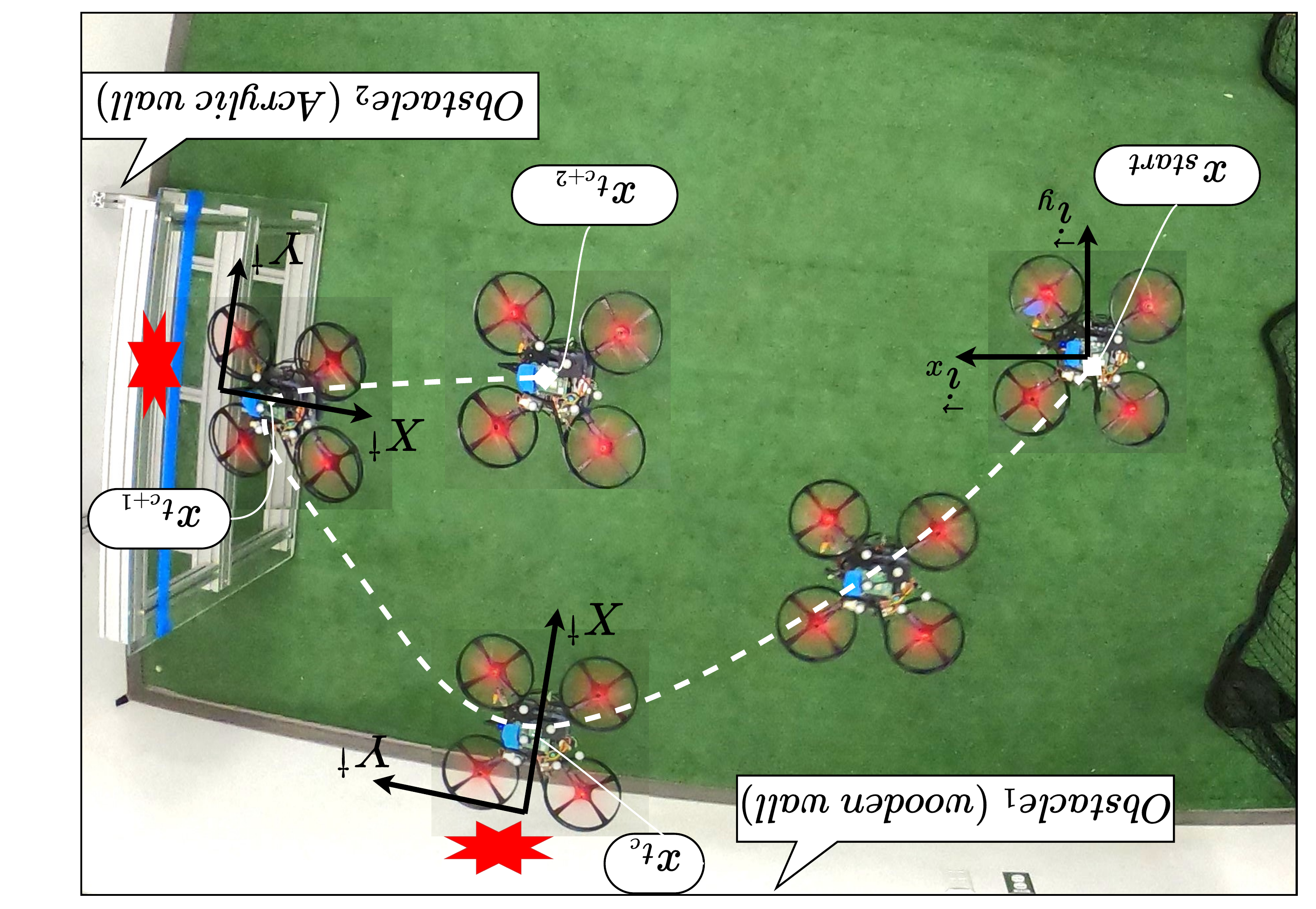}
    \caption{DART is a dual-mode variable stiffness quadrotor that adapts its post-collision response by switching between rigid and flexible modes. The figure demonstrates how this capability, along with a new collision response model, enables the generation of collision-inclusive paths.}
    \label{fig:first_pic}
    \vspace{-0.3in}
\end{figure}
Researchers have explored various methods for designing collision-resilient quadrotors, including making structural modifications and using different materials for the chassis \cite{KM+22}. Accordingly, quadrotors with compliant propeller guards/arms \cite{KB+13}, \cite{LK+21} and passive folding mechanism \cite{PM+20}, \cite{NP+23} have been explored to mitigate damage due to collisions. However, existing designs are typically single-mode: either safeguarded by propeller guards that prevent deformation or deformable but lacking rigidity, which is crucial for stable flight in open environments. As a result, it is a missing opportunity to develop quadrotors with variable body stiffness, as it allows for switching between compliant behaviors when absorbing impacts is beneficial and rigid behaviors when stability and higher rebound velocities are needed.

\begin{figure*}[t]
    \centering
    \includegraphics[trim = 0.7cm 0.6cm 0.65cm 0cm, clip, width=0.9\textwidth]{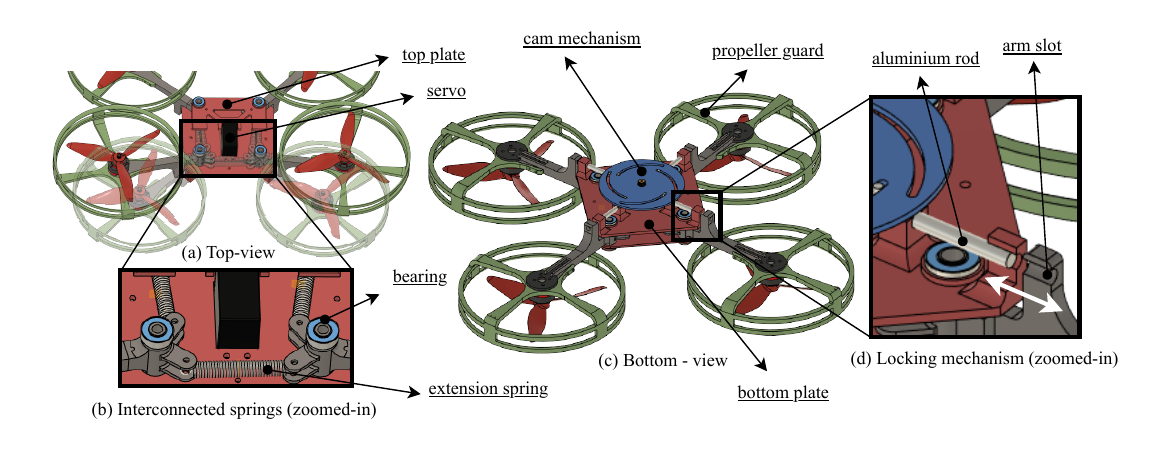}
    \caption{(a) Top-view (b) Arms mounted on bearings and interconnected using extension springs (c) Bottom-view (d) Locking mechanism}
    \label{fig:swift_CAD}
    \vspace{-0.2in}
\end{figure*}
Furthermore, existing efforts have been concentrated on mechanical design \cite{PZ21}, while understanding the drone-environment collisions and planning collision-inclusive trajectories have received little attention. As a result, most collision-resilient quadrotors employed manually-designed trajectories and focused on demonstrating the benefit of the compliant body design~\cite{NP+23,PS+23,PP+22,WP+23}. Few researchers have recently attempted to show the benefits of collision-inclusive trajectories \cite{ZL+23,LL+22,ZM21, MM18}. In some of these works, the post-collision behavior is modeled by assuming a reflected ray behavior which doesn't exploit the external force administered by the collision. 
Additionally, post-collision dynamics are typically modeled using Kelvin-Voigt models, which do not provide detailed estimates of contact forces \cite{AR+22}. In summary, there exist major research gaps due to the use of inaccurate collision models, inability to harness the collision energy and fully exploit the compliant designs, and lack of integration of recovery controllers within these algorithms.


In this paper, we introduce DART, a novel collision-resilient drone with dual stiffness that enables two distinct post-collision behaviors. By selecting two specific stiffness values, we demonstrate how the flexible mode facilitates smooth contacts and low-rebound velocities, while the rigid mode provides stability during free flight and enables maneuvers with higher rebound velocities. To better understand these behaviors, we develop a comprehensive collision model based on the theory of linear complementarity systems (LCS). We conduct drop and flight tests and use the post-collision data to identify and validate the LCS model. Furthermore, we show how this model can be used to select optimal collision modes and determine pre-collision states for waypoint tracking when intentional collisions are preferred. Finally, experimental results confirm DART's adherence to the predicted behavior and highlight its advantages in planning collision-inclusive trajectories, as illustrated in Fig. \ref{fig:first_pic}. In summary, the contributions of this work include
\begin{itemize}
    \item design of a novel dual-stiffness and collision-resilient quadrotor, DART,
    \item development of a novel 3D collision model to accurately capture and predict post-collision dynamics, and 
    \item demonstration of optimal mode sequence selection and pre-collision velocities for collision-inclusive trajectories using the collision model.
\end{itemize}

The rest of the article is organized as follows. Section \ref{sec:design} introduces the novel design to achieve the two different stiffness values. Section \ref{sec:characterization} details the collision modeling via various drop tests and flight tests. Section \ref{sec:LCS_overall} describes the LCS model to describe the collision-inclusive behavior with the recovery control. Finally, Section \ref{sec:results} presents the experimental results for the collision-inclusive path while Section \ref{sec:conclusion} concludes the article and presents future work.




\section{Design and Fabrication}\label{sec:design}

DART is built using custom parts via additive manufacturing techniques and commercially available electronic hardware. In this section, we will discuss the structural design, arm-locking mechanism, and drone assembly in detail. 

\subsection{Structural Design}
The design incorporates each arm mounted on bearings, allowing for free rotation, while extension springs are utilized to interconnect these arms within a specific configuration as shown in Figs. \ref{fig:swift_CAD}(a) and (b). The rotational degree of freedom about the pivot allows these arms to deform the chassis under the impact of external force/torque, and the extension springs provide an elastic nature to the overall structure, enabling the restoration of any deformation caused by external forces. The bearings with a 8mm inner diameter, 16mm outer diameter, and 5mm bore are used. The springs of the specification 1-7/8in length $\times$ 1/4in outer diameter $\times$ 0.054 water gauge are chosen after experimenting with various springs of different stiffness and length values. Our testing shows that the inclusion of springs with low stiffness induces vibrations during normal flight. Conversely, springs with high stiffness reduce the compliance of the arm to external impacts and are more likely to cause a breakage of the spring-mounting hook. Therefore, we choose springs of suitable stiffness to fulfill our requirement of moderate stiffness and minimum backlash after restoration.  
\subsection{Arm-locking Mechanism}

\begin{figure}[t]
    \centering
    \subfloat[Cam mechanism\label{fig:cam_mechanism}]{    \includegraphics[trim = 5.5cm 6.5cm 13.5cm 3cm, clip, width=0.24\textwidth]{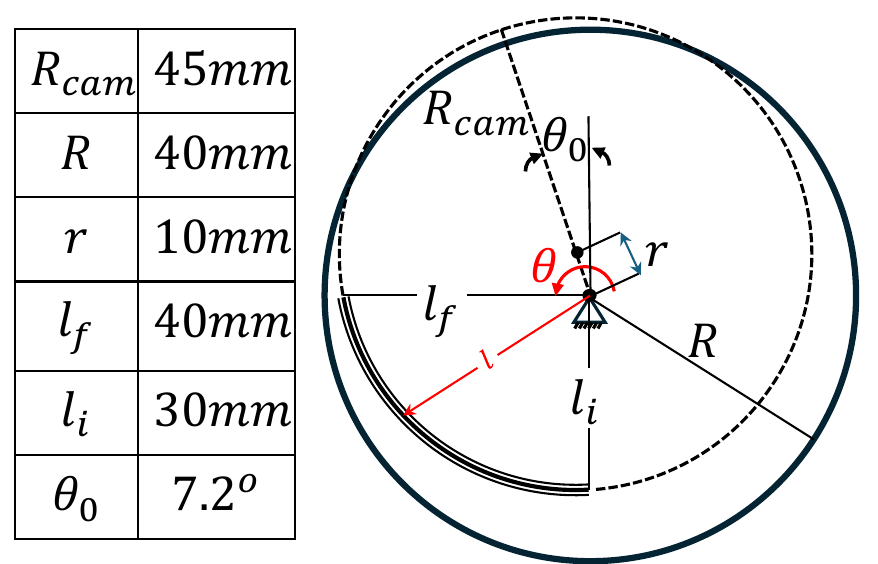}}
    \subfloat[Collision model \label{fig:motion_model}]{\includegraphics[trim = 4cm 0cm 4cm 0cm, clip,width=0.2\textwidth]{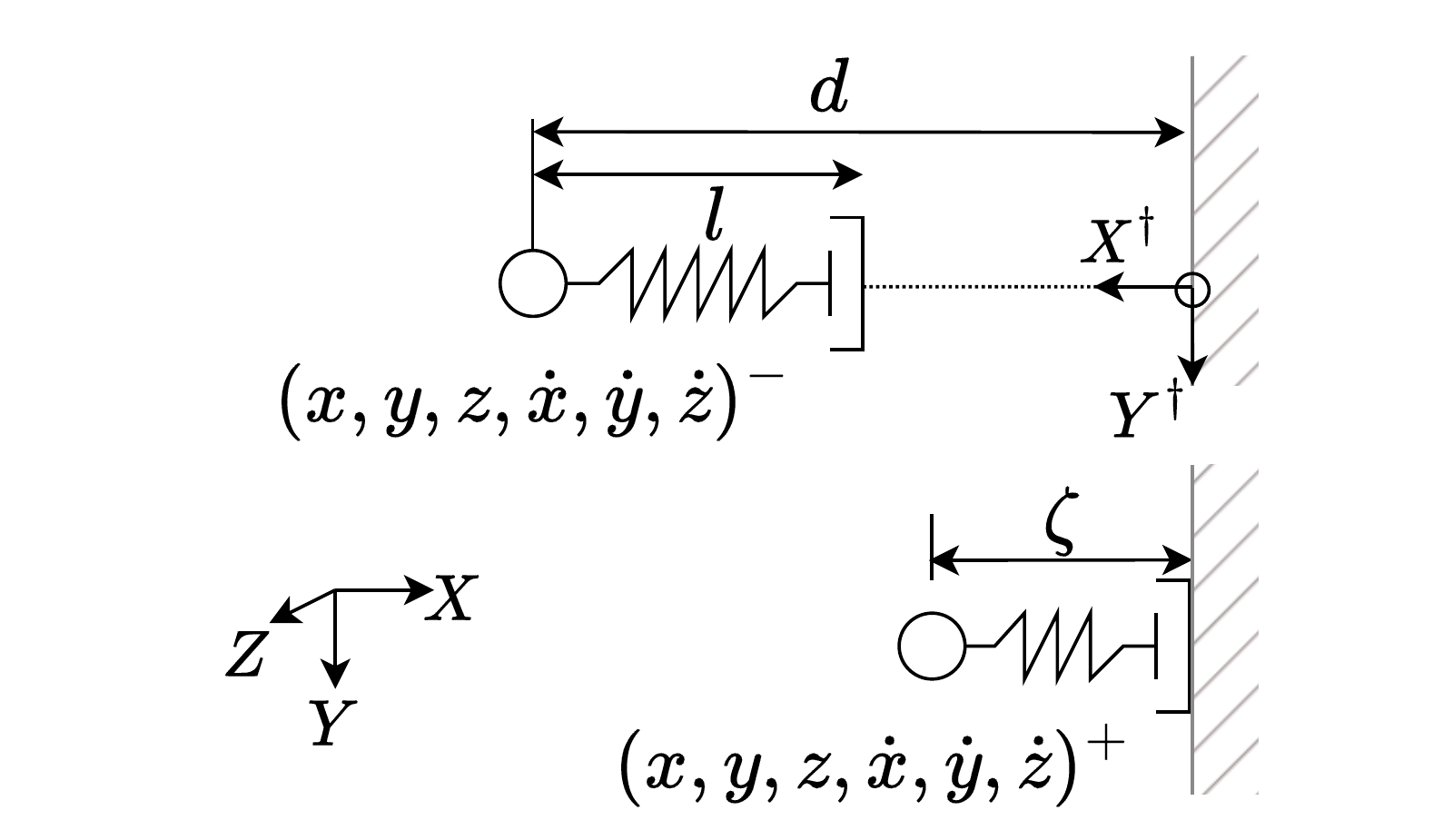}}
    \caption{(a) The kinematic model for the cam mechanism. (b) The motion model for DART, described as a 3D spring-damper system.}
    \label{fig:cam_and_model}
    \vspace{-0.2in}
\end{figure}

A locking mechanism as shown in Fig. \ref{fig:swift_CAD}(d) is used to lock/unlock the arms depending on the desired mode of stiffness to collide with the obstacle. The mechanism consists of aluminium rods (5mm diameter) that are attached to the four-face cam disc on one end and can fit into the arm slot on the other end. The rods are translated into slots by rotating the cam disc shown in Fig. \ref{fig:cam_mechanism} with kinematics model and parameters in (\ref{eq:1}). The locking mechanism is actuated using a servo motor to lock/unlock all four arms simultaneously. The locked configuration corresponds to a rigid mode and unlocked configuration refers to flexible mode of DART.

The relationship between $l$ (translation of aluminium rod) and $\theta$ (rotation of cam disc) is established as follows:
\begin{equation} \label{eq:1}
    l = R_{cam} - r \cos(\theta)
\end{equation}
where $R_{cam}$ and $r$ are cam and rotational circle radius respectively. For locking mechanism we define the rotational limits as $ \theta_0 \leq \theta \leq (\theta_0 + 90^o)$.

\subsection{Drone Components and Assembly}

The complete drone assembly includes various components fabricated from  Polylactide (PLA) material using additive manufacturing technology. These components consist of the arm, electronics mounting plates, propeller guards, cam disc, and landing gears. For the electronic components, the drone utilizes a commercial flight controller PX4 (Auterion, Zurich, Switzerland) and an onboard companion computer Raspberry Pi 4 (Raspberry Pi Foundation, Cambridge, England). Additionally, it incorporates four brushless DC motors Emax RS2205 paired with 6-inch propellers (Gemfan Flash 6042, Gemfanhobby Co. Ltd, Ningbo, China), along with a HS-322HD standard heavy-duty servo (HiTec RCD, USA).

\section{System Characterization}\label{sec:characterization}
This section provides a detailed description of the design characterization, including drop tests used to validate the variable stiffness achieved in the two modes. Our primary goal in conducting these drop tests was to develop a collision model based on LCS to characterize the post-collision behavior of DART. We chose drop tests over in-flight collisions because they allowed us to collect high-quality data with a high-speed camera, which was not feasible with in-flight experiments. Additionally, in-flight data can suffer from state estimation errors and detailed characterization of control inputs is needed. We work under the assumption that once we have a parametric collision model, we can fine-tune its parameters to characterize in-flight collision dynamics. 


Our experimental setup consists of an autonomous arm manipulator (UR5, Universal Robots, MI, USA), a high-speed camera operating at 1000 frames per second (Edgetronics SC1, CA, USA), and DART's chassis with added weight to match the total weight (1.3kg) of the assembled drone, including all electronic components and the battery. 
\begin{figure}[t]
    \centering
    \includegraphics[trim = .8cm .2cm 0.2cm .5cm, clip, width=0.45\textwidth]{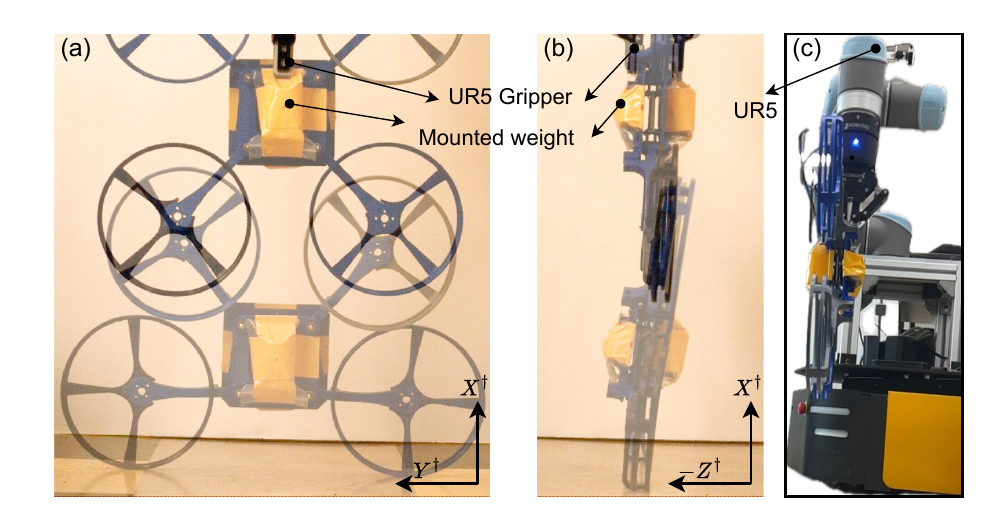}
    \caption{(a) Front-view and (b) Side-view, showing initial and maximum flexed configurations of DART during the drop test from 0.2m height (c) UR5 holding DART for drop test}
    \label{fig:droptest_front_side_view}
    \vspace{-0.2in}
\end{figure}
In a drop test, the drone is securely mounted on the arm manipulator using the gripper and released from a predetermined height. The entire drop process as shown in Fig. \ref{fig:droptest_front_side_view} is recorded using the high-speed camera. We examine four distinct configurations for the experiment: dropping with unlocked and locked arms from two different heights, 0.05m and 0.2m to resemble the collisions with 1m/s and 2m/s velocities respectively. The captured video for each drop is processed using Tracker (a free video analysis and modeling tool, a project of Open Source Physics foundation). 

\begin{figure}[t]
    \centering
    \subfloat{\includegraphics[width = 0.24\textwidth]{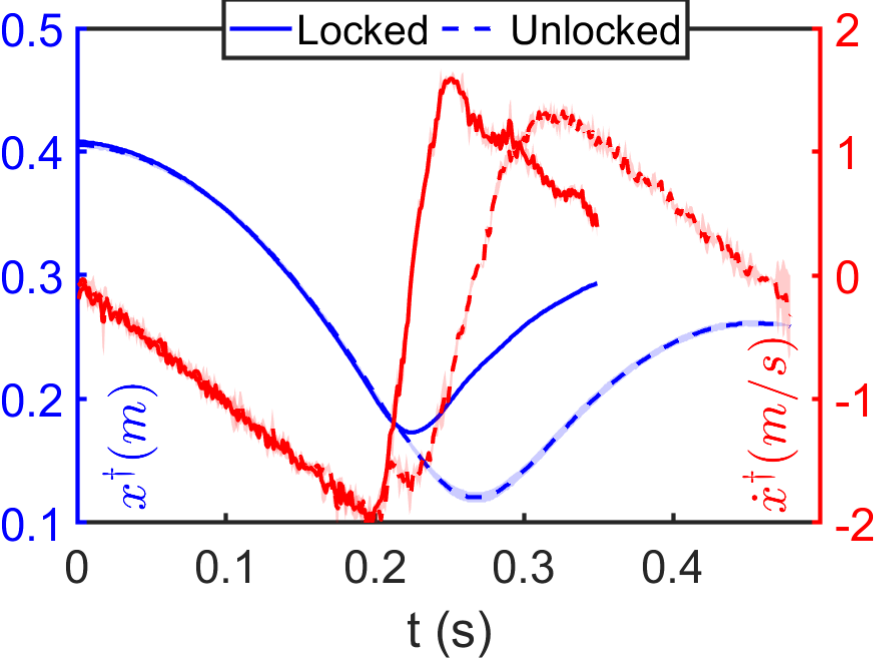}} 
    \subfloat{\includegraphics[width = 0.24\textwidth]{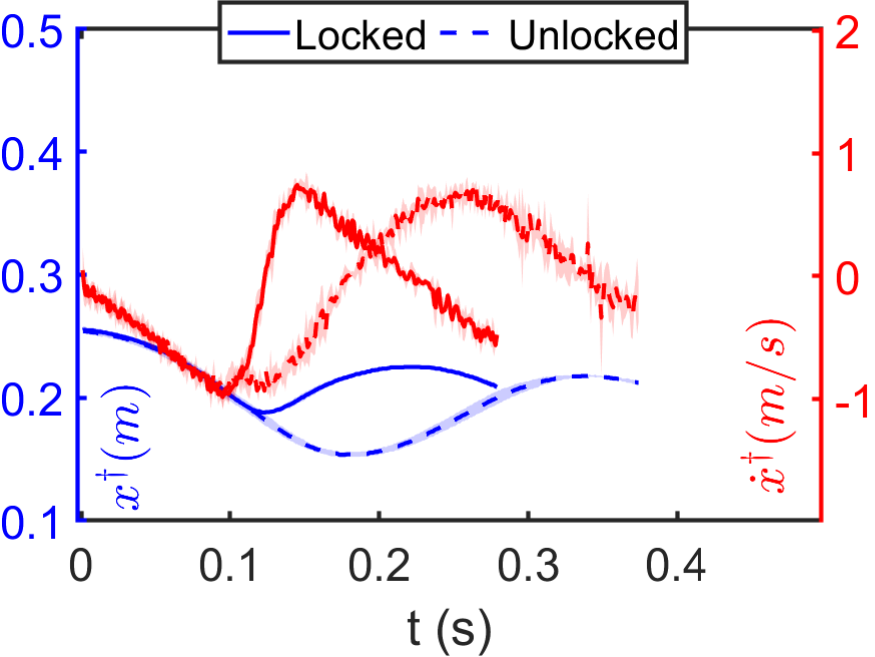}} \\ \vspace{-0.15in}
    \subfloat{\includegraphics[width = 0.24\textwidth]{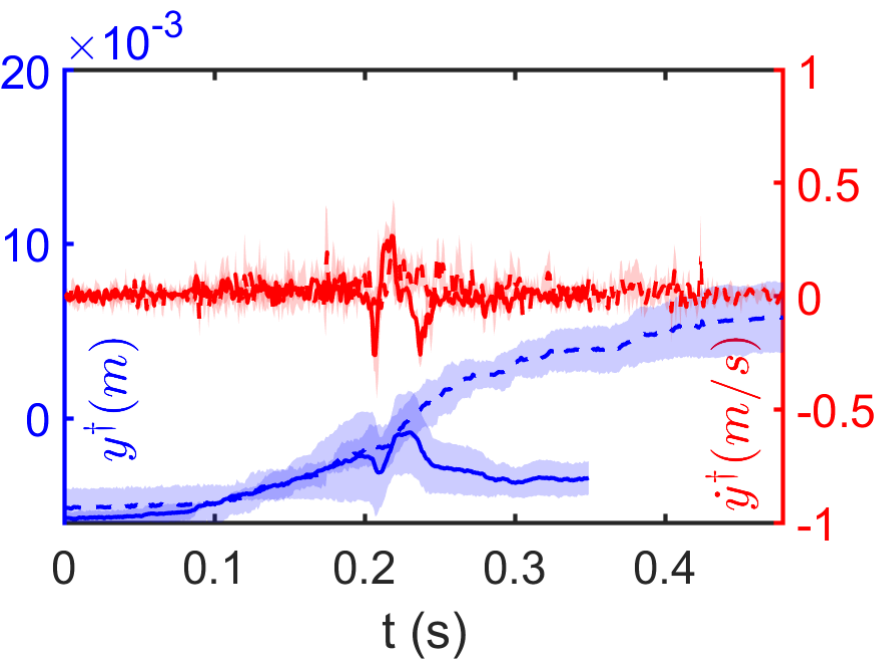}}
    \subfloat{\includegraphics[width = 0.24\textwidth]{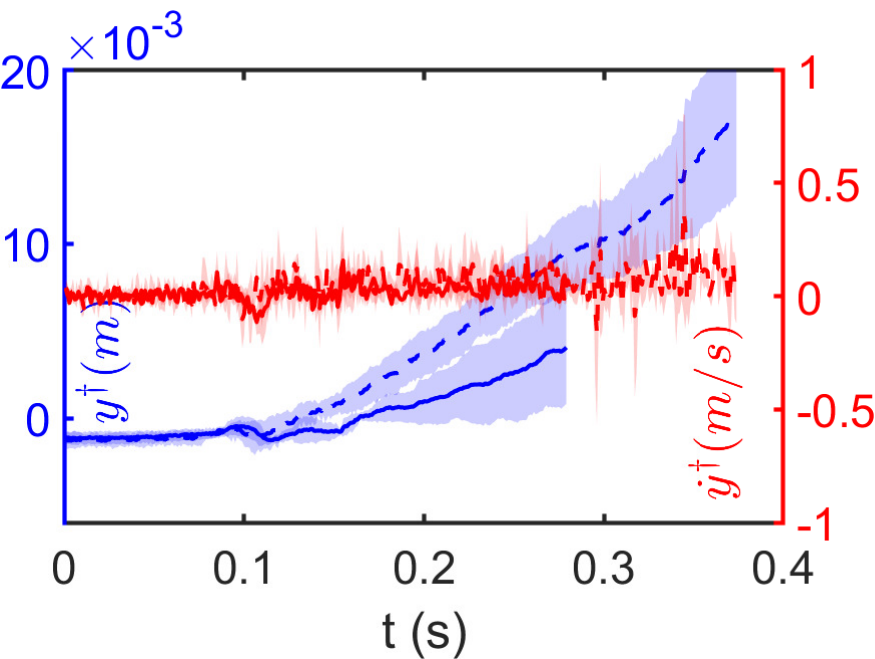}} \\ \vspace{-0.15in}
    \subfloat{\includegraphics[width = 0.24\textwidth]{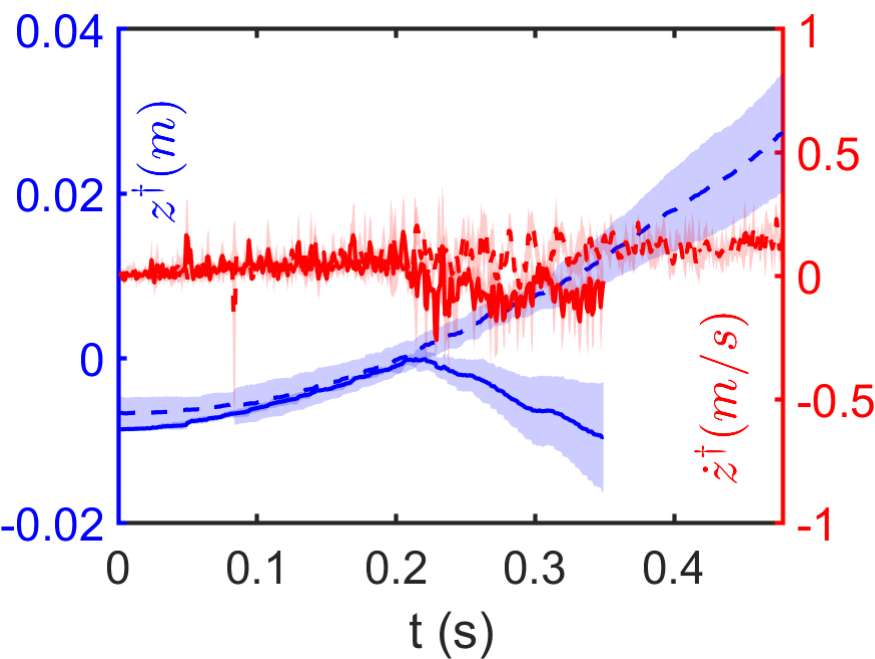}} 
    \subfloat{\includegraphics[width = 0.24\textwidth]{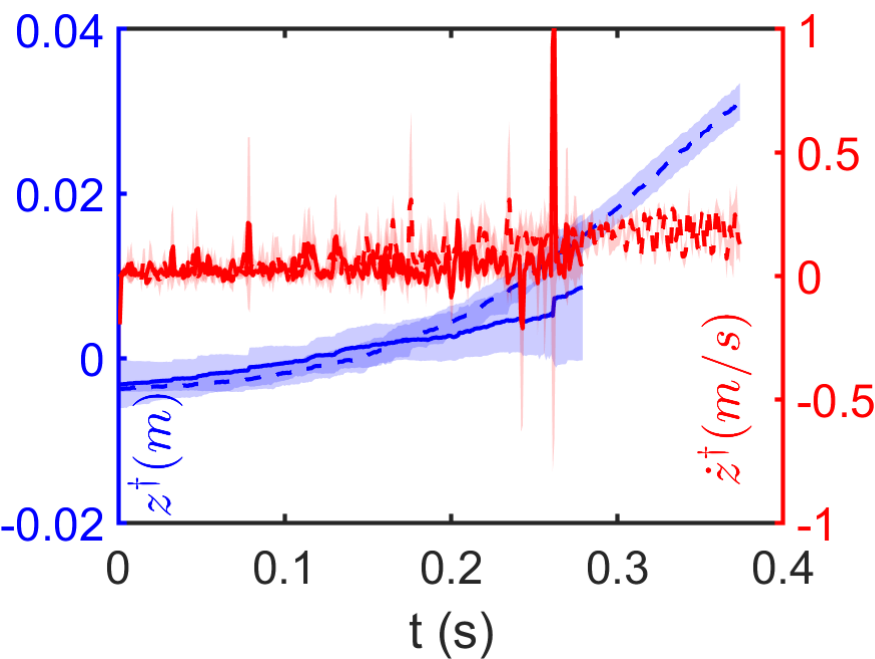}}
    \caption{Position ($x^\dagger$) and velocity ($\dot{x}^\dagger$) vs time from drop test experiments}
    \label{fig:drop test}
    \vspace{-0.3in}
\end{figure}

The results of the experiment are depicted in Fig. \ref{fig:drop test}. The ($\boldsymbol{x}^\dagger, \dot{\boldsymbol{x}}^\dagger$) denoting position and velocity of the center of mass are tracked, which is 0.2m high from the ground when first contact is made. The $x^\dagger$ (m) values plotted on the left $y$-axis represent the height from the ground, revealing a noticeable quick bounce-back in case of rigid mode compare to the flexible mode for both 0.05m and 0.2m drop heights. The contact time for rigid and flexible mode is 46ms and 177ms for the 0.05m drop height, and 52ms and 145ms for the 0.2m drop height, respectively. The maximum bounce-back velocity observed for the 0.05m drop height is 0.74m/s and 0.70m/s for the rigid and flexible configurations, respectively. For the 0.2m drop, the maximum velocities are 1.59m/s and 1.34m/s for the rigid and flexible configurations, respectively. From these results, we can conclude that flexible mode has significantly larger contact time and it reduces the effect of impact force on chassis, compared to the rigid mode. 


\section{Collision Model using Linear Complementarity Systems}\label{sec:LCS_overall}
In this section we present a novel collision model for any generic collision-resilient quadrotor based on the theory of linear complementarity systems. We validate the model for DART, and demonstrate its significance for generating collision-inclusive trajectories. We start this section by describing the collision recovery controller for guaranteeing stability during collisions and the low-level flight controller, followed by the LCS model for collision contact force prediction and the planning method which leverages this collision model.

\subsection{Description of Recovery Controller}\label{sec:recovery_ctrl}
\begin{figure}[t]
    \centering
    \includegraphics[trim = 0.7cm 0cm 0.7cm 0cm, clip, width = 0.48\textwidth]{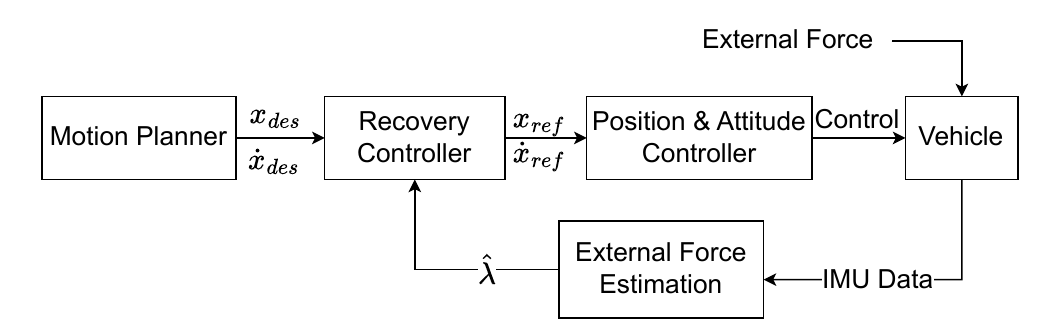}
    \caption{Control block diagram for a collision-inclusive maneuver, by utilizing a recovery controller.}
    \label{fig:cbd}
    \vspace{-0.2in}
\end{figure}

In our collision inclusive framework, we incorporate a recovery controller that uses the external force estimation to generate a post-collision set-point in the direction of the external force and generated according to a proportionality gain. The structure of the controller is shown in Fig. \ref{fig:cbd}. The recovery controller is based on only acceleration-based techniques for ease of implementation \cite{PM+21}.

During a collision-inclusive maneuver the quadrotor detects the collision if force estimator detects a reading above the threshold ($\lambda_{th}$). We consider collision-inclusive path planning in the hovering plane ($x$-$y$) only. Hence, if estimated forces in inertial frame $\hat{\lambda}_x ~or ~\hat{\lambda}_y \geq \lambda_{th}$ at impact time ($t_c$), a collision is registered. We generate the post-collision setpoints $x_{ref}(t_{c+1})$ and $y_{ref}(t_{c+1})$ given the instantaneous $\hat{\lambda}_x(t_c)$ and $\hat{\lambda}_y(t_c)$ as (\ref{eq:2}) with a constant $z_{ref}(t)$.
\begin{equation}
\begin{aligned} \label{eq:2}
    x_{ref}(t_{c+1}) = x(t_c) + \alpha_x \hat{\lambda}_{x}(t_c) \\
    y_{ref}(t_{c+1}) = y(t_c) - \alpha_y \hat{\lambda}_{y}(t_c)
\end{aligned}
\end{equation}
The constants $\alpha_x$ and $\alpha_y$ are parameters that can be tuned based on the desired post-collision behavior for every collision. $(x(t_c)$, $y(t_c))$ represents the vehicle position at the collision instant.

For the low-level controller, we use the classic P-PID control structure of the PX4 Autopilot. The P-PID architecture is seen to work efficiently since the arm deformation happens very fast and only for a short time with zero backlash in the springs. This ensures that the drone does not destabilize during and after the collisions.

\subsection{LCS Formulation for Collision Model}\label{sec:LCS}
This section describes the LCS system developed to model DART's behavior before and after collisions, along with the key assumptions made during the model's derivation. 

In this model, the angle of incidence—defined as the angle between DART's leading edge and the collision plane—is assumed to be zero. This assumption is supported by the deformation behavior observed during impact, as shown in Fig. \ref{fig:first_pic}, where both propeller guards compress. The deformation effectively transforms any two-point collision into a linear spring-damper system, justifying the zero-angle approximation. Furthermore, it is observed during experiments that when DART hits the collision surface with significant angle of incidence (i.e, head-on collision), the collision model $\Gamma = 1$ can be used to predict the response, since even with the arms unlocked, it behaves like a rigid counterpart.

We model DART as a point-mass with a unilateral spring (that can only apply pushing force) and damper system attached to its colliding end, as shown in Fig. \ref{fig:motion_model}. Furthermore, we represent the collision mode of DART by $\Gamma \in \{1,2\}$ indicating either the rigid or flexible mode, respectively.
Now, let $x,y,z$ denote the position of the vehicle in the world inertial frame $\begin{Bmatrix}
\vec{i}_{x},\vec{i}_{y},\vec{i}_{z} \end{Bmatrix}$, $\dot{x}, \dot{y}, \dot{z}$ denote the translational velocities in the world frame, and $b_x,b_y,b_z$ denote the air drag in $\vec{i}_{x}$, $\vec{i}_{y}$ and $\vec{i}_{z}$ directions, respectively. We now introduce the collision frame $\{X^\dagger~ Y^\dagger ~Z^\dagger\}$  with its origin at the intersection of the collision plane and the line of motion and define $(\cdot)^\dagger$ as the quantity $(\cdot)$ in the collision frame. In this frame, $X^\dagger$ points towards the outward normal at the origin, and $Y^\dagger$ points to the right, with the remaining axis following the right-hand rule. The collision frame is used to simplify the contact dynamics simulations and $x^\dagger$ will represent the signed distance between the point mass and the collision plane. The collision frame is illustrated in the drop tests of Fig. \ref{fig:droptest_front_side_view}.


Since quadrotors are differentially flat systems with the four inputs of 3D position and yaw angle, a 3D point mass is used to develop the translational motion model and yaw is chosen such that the drone is parallel to the colliding edge. Hence in vector form, let $\boldsymbol{x} =[x,y,z]^T\in \mathbb{R}^{3} $,  $\boldsymbol{u}=[u_x,u_y,u_z]^T \in \mathbb{R}^{3}$ and $\boldsymbol{b}=[b_x,b_y,b_z]^T \in \mathbb{R}^{3}$, the 3D translational dynamics can be then written as
\begin{equation}
    \Ddot{\boldsymbol{x}} = g(\boldsymbol{u}) + diag(\boldsymbol{b})\boldsymbol{\dot{x}}.
\end{equation}

To predict the post-collision behavior, we model this point mass with a spring-damper system as a LCS in the state space form with the state vector $\boldsymbol{\bar{x}} =[\boldsymbol{x}^{T},\boldsymbol{\dot{x}}^{T}]^T\in \mathbb{R}^{6}$ and velocity input vector $\boldsymbol{\bar{u}} = [\boldsymbol{u}^{T}, \mathbf{0}^{T}_{3}]^T\in \mathbb{R}^{6}$
\begin{equation}
    \boldsymbol{\dot{\bar{x}}} = \boldsymbol{A}\boldsymbol{\bar{x}} + \boldsymbol{B\bar{u}} + \boldsymbol{\lambda} 
\end{equation}
where $\boldsymbol{\lambda}  = [\mathbf{0}^{T}_{3},\boldsymbol{R}^\dagger\boldsymbol{\bar{\lambda}}^{T}]\in \mathbb{R}^{6}$ denotes the contact force and $\boldsymbol{R}^\dagger \in \mathbb{R}^{3\times3}$ denotes the rotation matrix from colliding reference frame to inertial frame. The vector $\boldsymbol{\bar{\lambda}} = [\lambda_{x^\dagger} ,\lambda_{y^\dagger},\lambda_{z^\dagger}]$ denotes the 3D contact force, in the colliding reference frame, as experienced by the point mass during contact. 

\begin{figure*}
    \centering
    \subfloat[0.05m locked]{\includegraphics[width = 0.25\textwidth]{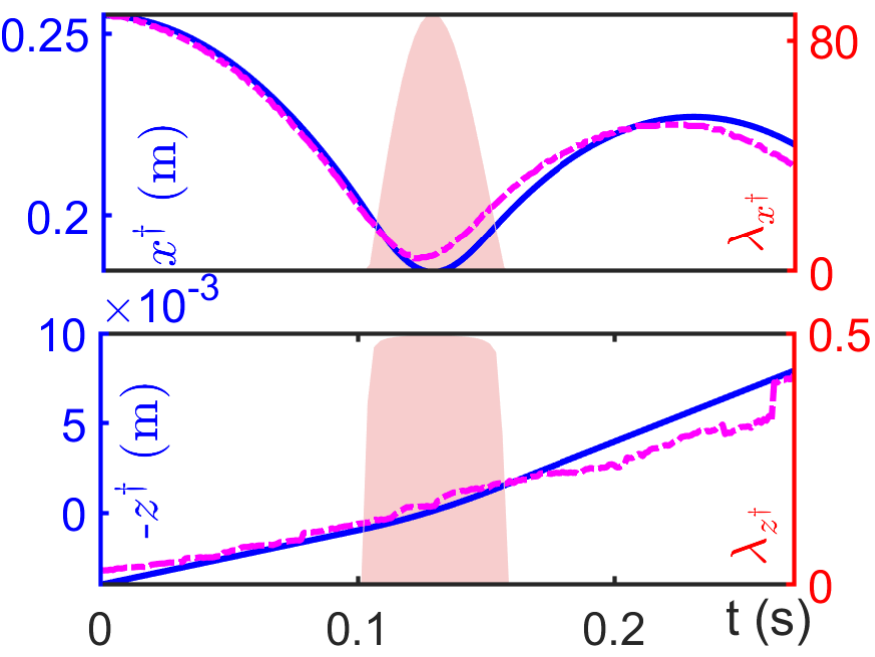}}
    \subfloat[0.05m unlocked]{\includegraphics[width = 0.25\textwidth]{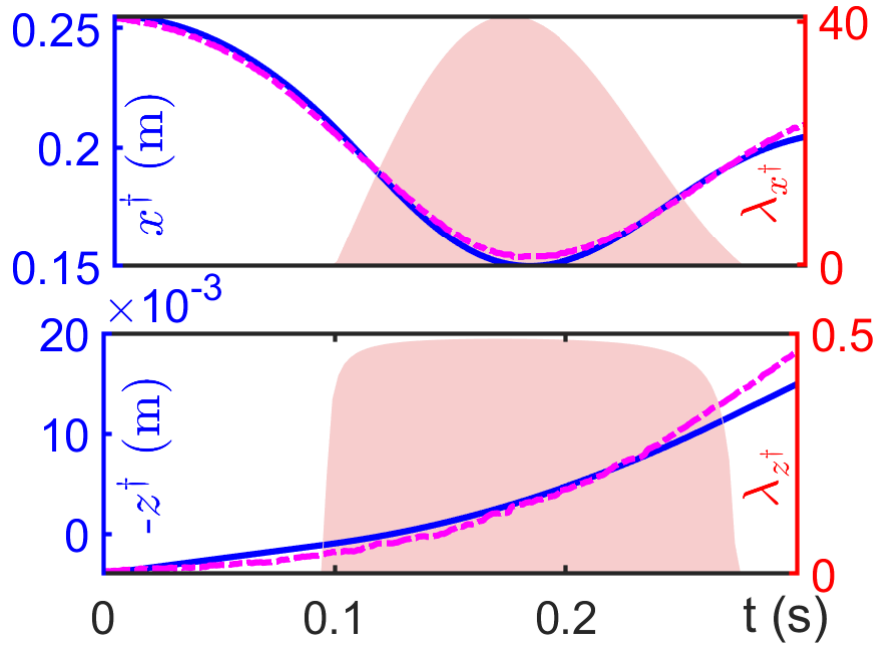}}
    \subfloat[0.2m locked]{\includegraphics[width = 0.25\textwidth]{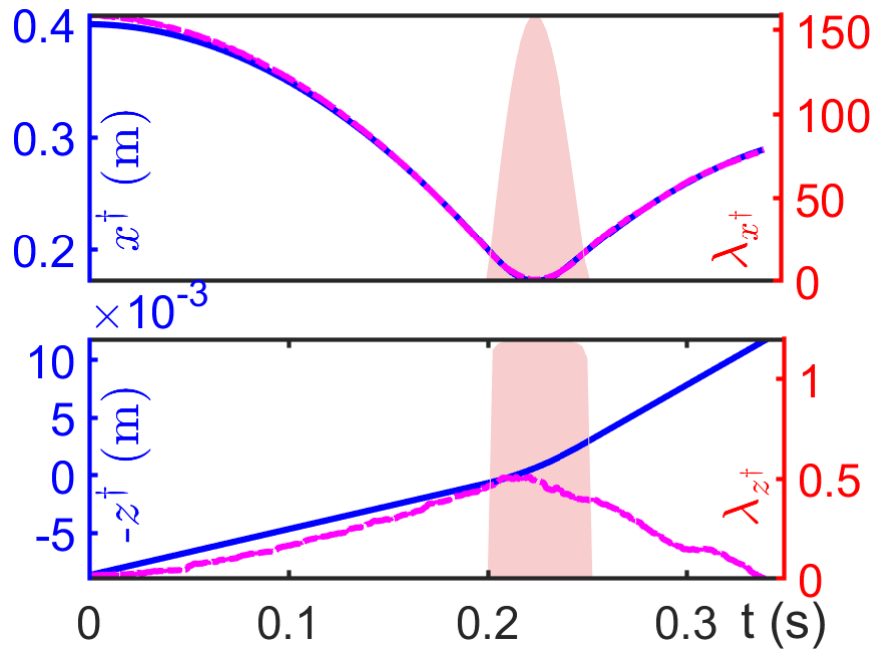}}
    \subfloat[0.2m unlocked]{\includegraphics[width = 0.25\textwidth]{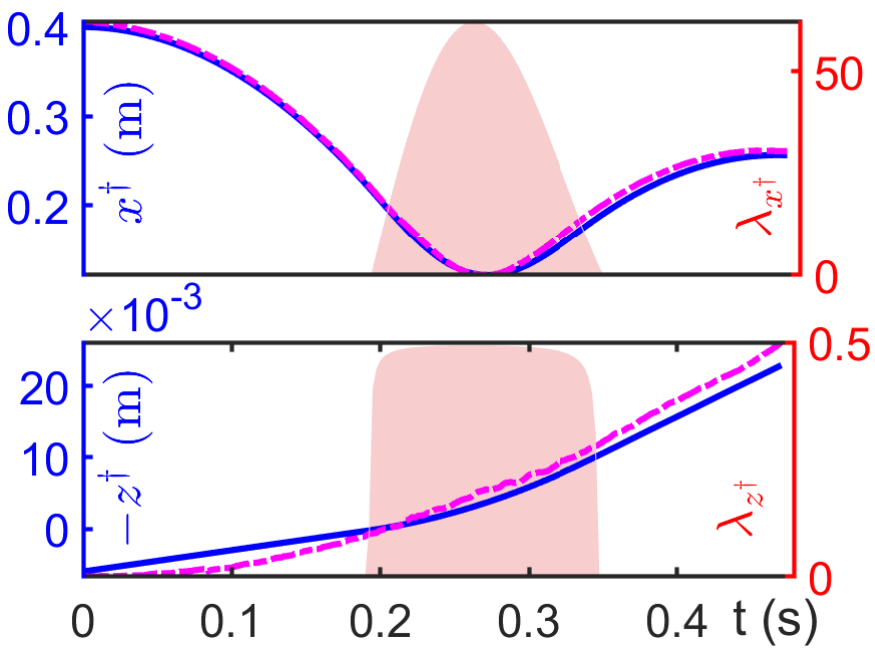}}
    \caption{Validation results of the LCS-based collision models with drop test data of DART's two modes. The dash-dotted magenta lines represent the experimental data, while the solid blue lines indicate the model fit. The shaded region plots the contact force $\lambda_{x^\dagger}$ and $\lambda_{z^\dagger}$ predicted during the collision duration. $\lambda_{y^\dagger}$ is seen to be negligible due to negligible $\dot{y}^\dagger$, hence is not plotted here. Note the experiments denote the average over 3 trials. }
    \label{fig:droptest_validation}
    \vspace{-0.2in}
\end{figure*}
We use the LCS theory to estimate $\boldsymbol{\bar{\lambda}}$. The contact force $\lambda_{x^\dagger}$ is assumed perpendicular to the colliding plane and is the solution to the the following Lemke's problem \cite{L78}:
\begin{equation}
    0 \leq \lambda_{x^\dagger} \perp w(x^\dagger,\lambda_{x^\dagger}) = \lambda_{x^\dagger} + k x^\dagger + f \dot{x}^\dagger \geq 0
\end{equation}
with $k$ and $f$ denoting the spring and damping coefficients for the system. Furthermore, the frictional forces $\lambda_{y^\dagger}$ and $\lambda_{z^\dagger}$ are found to have the following form as functions of the contact force $\lambda_{x^\dagger}$ with dimensionless constants $\mu$ and $\nu$ :
\begin{equation}
    \begin{aligned}
        \lambda_{y^\dagger} = - \mu \lambda_{x^\dagger} \dot{y}^\dagger \\
        \lambda_{z^\dagger} = - \nu \frac{\lambda_{x^\dagger}}{1 + \lambda_{x^\dagger}}
    \end{aligned}
\end{equation}

We use \textit{implicit-Euler} method to solve for the contact dynamics and estimate the contact force \cite{ST99,B99}. Data from the 0.05m drop is employed to identify the coefficient $(k,f,\mu,\nu )$ of the collision model as was described in Section \ref{sec:characterization}. The collision model coefficients for the two different modes of rigid/locked ($\Gamma = 1$) and flexible/unlocked ($\Gamma = 2$) against a concrete ground are approximately:
\begin{align}
    (k,f,\mu,\nu )_{\Gamma = 1} = (5500\text{N/m}, 15\text{N-s/m}, 0.3, 0.5) \\
    (k,f,\mu,\nu )_{\Gamma = 2} = (750\text{N/m}, 8.5\text{N-s/m}, 5, 0.7) 
\end{align}
The model fitting data 
is shown in Figs. \ref{fig:droptest_validation}(a-b) while the validation results for the 0.2m drop are shown in Figs. \ref{fig:droptest_validation}(c-d). The collision model performs very well in predicting the post-collision state for the 0.2m drop. However, there is clear discrepancy in predicted value for the $z^\dagger$ direction contact force for the locked case in the orders of 0.001, shown in Fig. \ref{fig:droptest_validation}(c). This can be attributed to unmodeled moments about the impact point. Since in real flight tests, the flight controller is engaged to maintain constant height. This discrepancy will be addressed in future work.


\subsection{Collision Model Applications}\label{sec:CIT}
Here, we briefly describe the applications of the proposed collision model to generate collision inclusive trajectories. For the simulations, $b_x = 0.2$ and $b_y = 0.4$ to model the response in free flight  at constant height. 

\subsubsection{Determining Pre-collision Velocity}\label{sec:pre-collision_vel}
In this case study, we recreate the classical example from optimal control theory where it is required to move from start to goal point by visiting a plane. We employ exhaustive enumeration to find the optimal pre-collision or impact velocity and the corresponding collision mode. Towards this, forward simulations are performed to find the ideal pre-collision velocities from a given range of $\dot{x}$ and $\dot{y}$ which combined with collision model and the recovery controller would lead to predicted post-collision states. Again, the trajectory whose terminal state is closest to the goal state is chosen with the corresponding velocity as the desired pre-collision velocity. The simulation results for a single case are shown in Fig. \ref{fig:precollision_state}. Experimental results are discussed in Section \ref{sec:results}.

\subsubsection{Mode Selection}\label{sec:mode_selection}
\begin{figure}
    \centering
    \subfloat[\label{fig:precollision_state}]{\includegraphics[trim = 0 0 0cm 0cm, clip, width=0.5\linewidth]{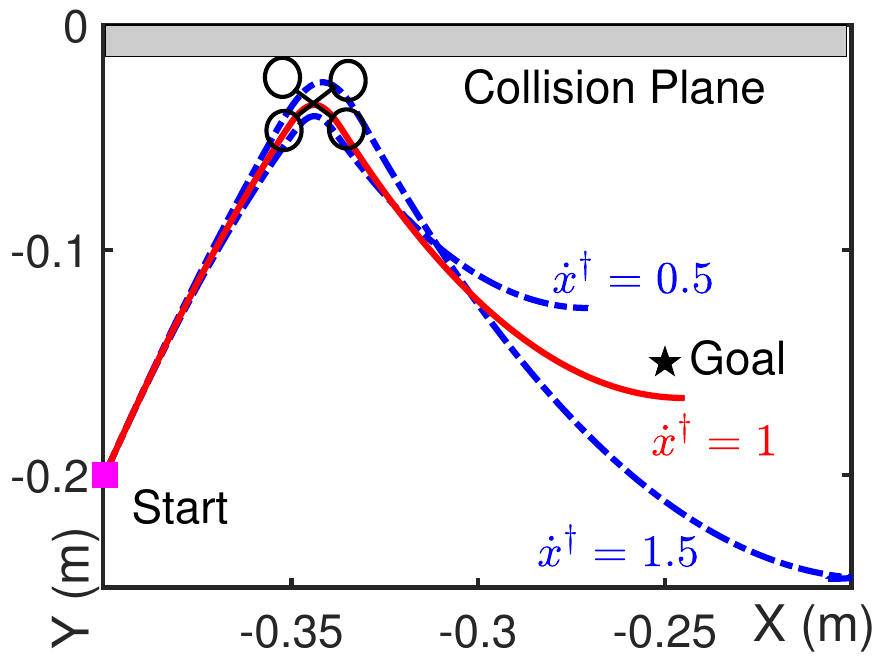}}
    \subfloat[\label{fig:mode_selection}]{\includegraphics[trim = 0 0 1cm 0.5cm, clip, width=0.5\linewidth]{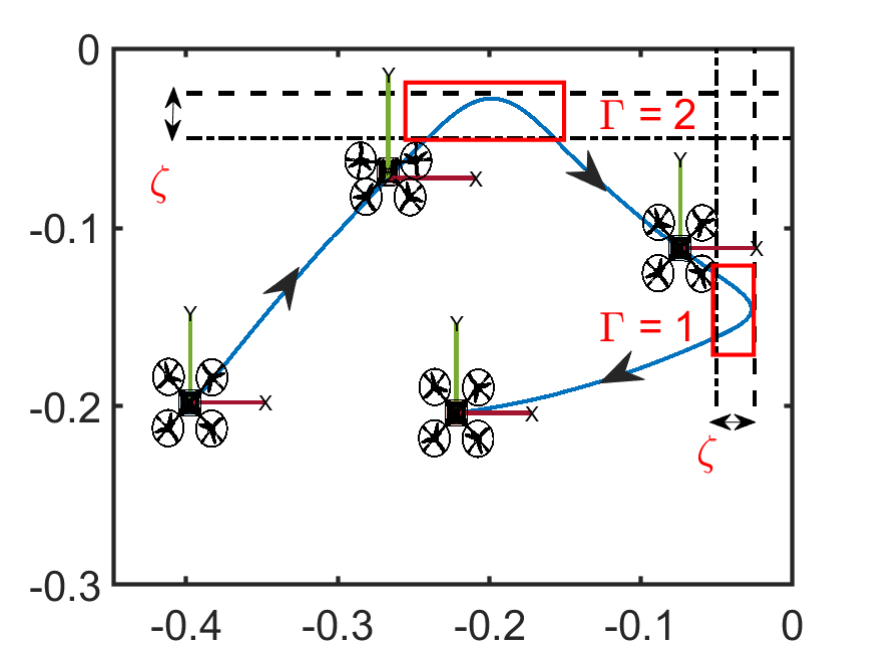}}
    \caption{Simulation results for (a) optimal \textit{collision velocity} determination by simulating responses across different approach velocities and selecting the most suitable one  and (b) optimal \textit{mode selection} by forward simulating various responses and choosing the best mode sequence}
    \label{fig:usecases}
    \vspace{-0.2in}
\end{figure}

The second use case we demonstrate is the selection of the optimal collision mode sequence by simulating trajectories for the different pre-collision velocities. We first consider the start and goal locations as shown in Fig. \ref{fig:mode_selection} by the magenta square and the black star markings. First, all the possible trajectories are computed by forward simulations considering various impact velocities and mode sequences. Considering the flight space in our lab, the pre-collision velocities were limited to $\pm$1.5m/s for both $\dot{x}^\dagger$ and $\dot{y}^\dagger$.
After the forward propagation, the trajectory with the nearest terminal value to the goal is selected and the corresponding mode sequence is chosen as the optimal one. Since we have discretized approach velocities in two mode, exhaustive enumeration effectively finds a solution for this problem. We demonstrate the experimental results for this case in Section \ref{sec:results} by employing the recovery controller from Section \ref{sec:recovery_ctrl}.

\section{Experimental Results}\label{sec:results}
This section presents the experimental results of the collision model validation and its integration for executing collision-inclusive trajectories. Our experimental setup, shown in Fig. \ref{fig:first_pic}, includes two obstacles: (i) a wooden wall and (ii) an acrylic wall for collision tests. To generate post-collision setpoints, we set the parameters ($\alpha_x,\alpha_y)$ in (2) to (0.12, 0.025) for collision 1 and (0.025, 0.025) for collision 2.  These values were fine-tuned based on preliminary testing to achieve the desired post-collision response.
\vspace{-0.1in}
\subsection{Collision Model Validation In-flight}
The collision model derived from the drop tests was validated through flight experiments involving impacts against an acrylic wall, where the maximum accelerometer values were recorded from the onboard PX4 flight controller. As expected, the stiffness, damping, and friction coefficients vary depending on the material properties of the colliding surface. Consequently, we had to adjust the spring coefficients obtained from the drop test model. The updated coefficients for DART's collisions with the acrylic wall are:
\begin{align}
    (k,f,\mu,\nu )_{\Gamma = 1} = (4200\text{N/m}, 15\text{N-s/m}, 0.3, 5) \\
    (k,f,\mu,\nu )_{\Gamma = 2} = (1250\text{N/m}, 8.5\text{N-s/m}, 5, 10) 
\end{align}
Interestingly, the spring coefficient for the rigid mode with $\Gamma = 1$ decreased from 5500N/m to 4200N/m. In contrast, the coefficient for the flexible mode, $\Gamma = 2$ increased from 750N/m to 1250N/m. This shift can be attributed to the compliance introduced by the imperfect mounting of the acrylic wall, which, despite our best efforts, did not replicate the rigidity provided by the ground.
\begin{figure}
    \centering
    \subfloat[$\Gamma = 1$]{\includegraphics[width=0.5\linewidth]{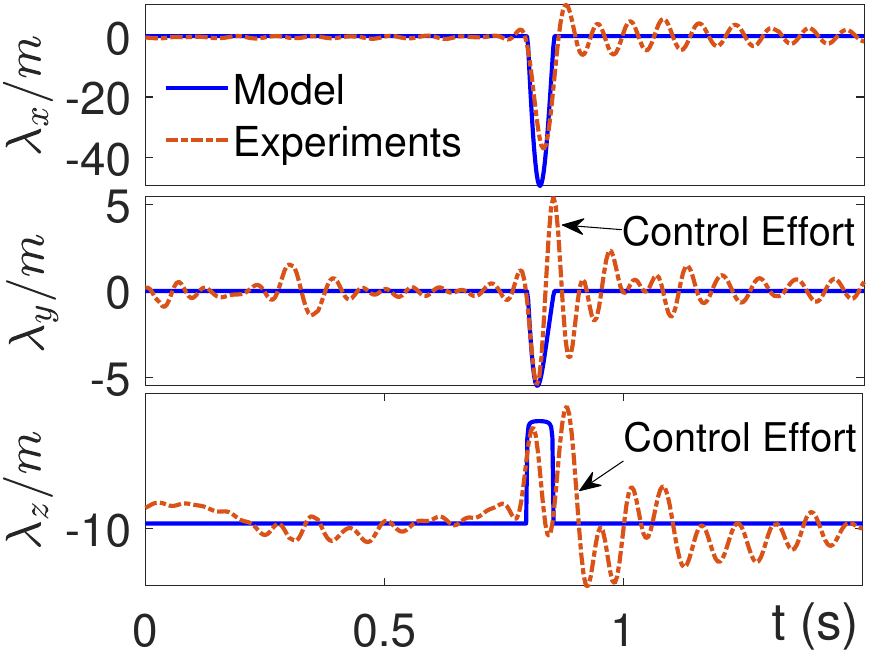}}
    \subfloat[$\Gamma = 2$]{\includegraphics[width=0.5\linewidth]{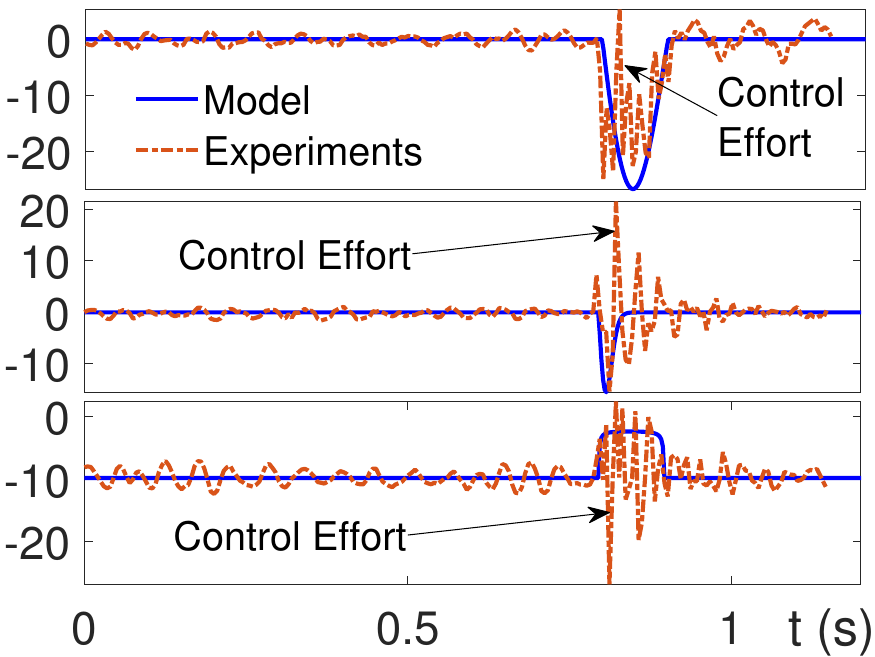}}
    \caption{Collision model predictions of peak contact force trends (inertial frame) for (a) rigid and (b) flexible modes for in-flight collisions at $\dot{x}^{-}$ = 1m/s and $\dot{y}^{-}$ = 0.5m/s. The second high peaks correspond to the control efforts due to the engagement of the recovery controller and damped oscillations due to the arm vibration. }
    \label{fig:inflight_validation}
    \vspace{-0.2in}
\end{figure}

The model validation results for collision flights against an acrylic wall, shown in Fig. \ref{fig:inflight_validation}, closely align with the collision characteristics recorded by the onboard inertial measurement unit (IMU) of the PX4 flight controller. To validate the peak contact force, we converted it into peak acceleration with $m = 1.3$kg and compare it against the IMU data. The results were consistent across three trials for each mode, showing nearly 88\% accuracy for the peak acceleration values in the flexible configuration and 82\% accuracy in the rigid mode for both $\lambda_{x}$ and $\lambda_{y}$. The $\lambda_{z}$ is seen to be vulnerable to the impacts, sometimes registering very high peaks, though the average peak value over the entire contact duration remained within $\pm$10\% of the predicted value. It's important to note that some post-impact peaks may also be caused by the recovery controller, which engages immediately after detecting a peak. Additionally, the damping oscillations after collisions are attributed to slight in-plane arm vibrations due to the flexible nature of the chassis. 



\begin{figure}
    \centering
    \vspace{-0.02in}
    \subfloat{\includegraphics[trim = 0.5cm 0.5cm 1.5cm 0.5cm,clip,width=0.24\textwidth]{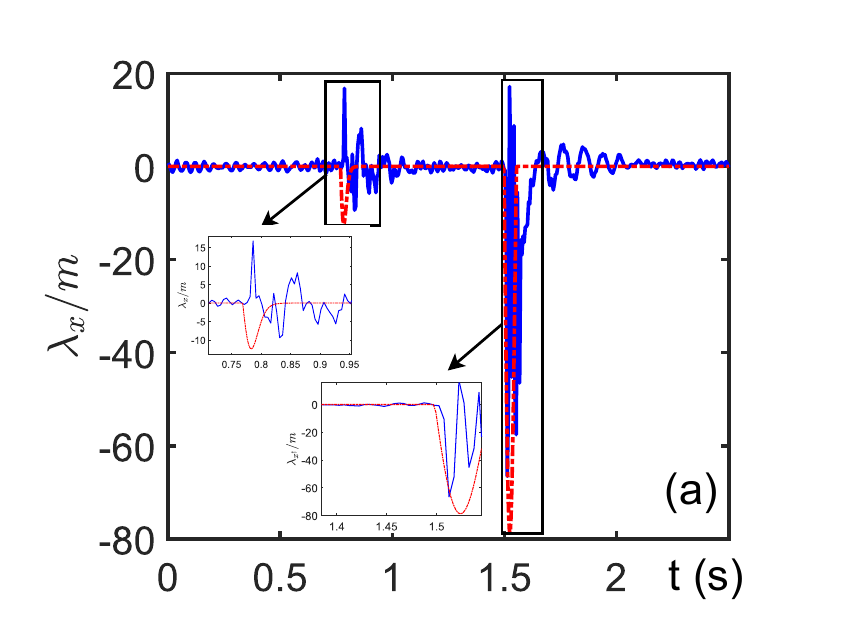}}
\subfloat{\includegraphics[trim = 0.5cm 0.5cm 1.5cm 0.5cm,clip,width=0.24\textwidth]{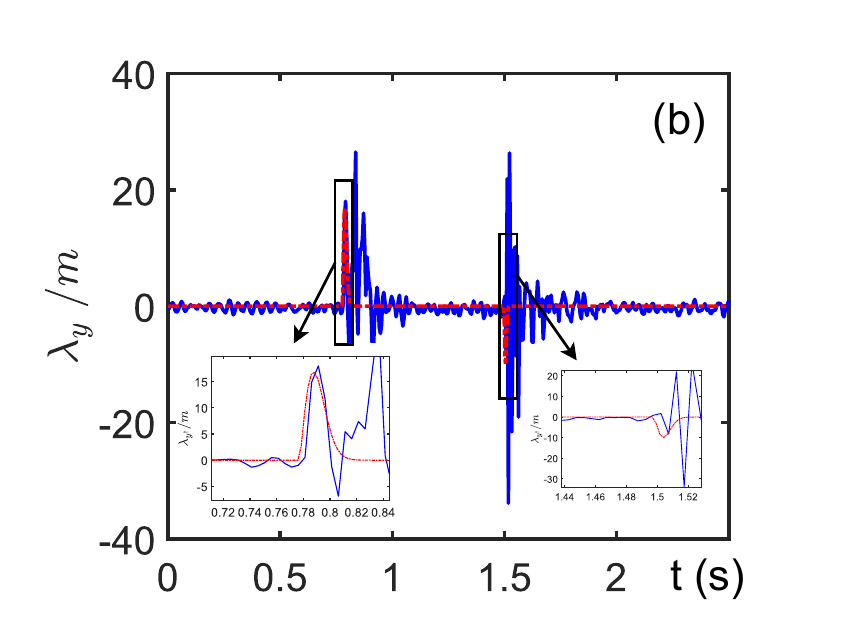}} \\
    \vspace{-0.02in}
\subfloat{\includegraphics[trim = 0.5cm 1.15cm 1cm 0.6cm,clip,width=0.24\textwidth]{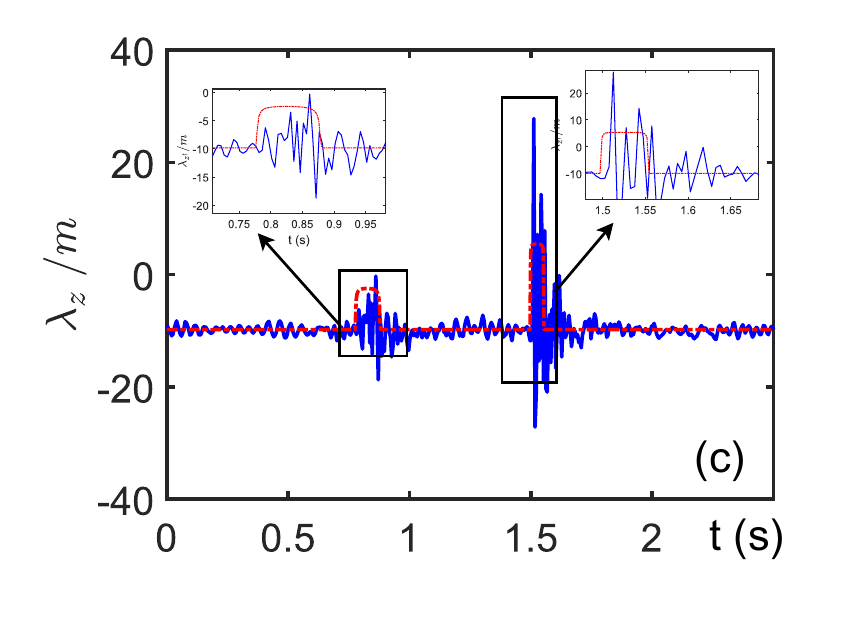}}
\subfloat{\includegraphics[width=0.23\textwidth]{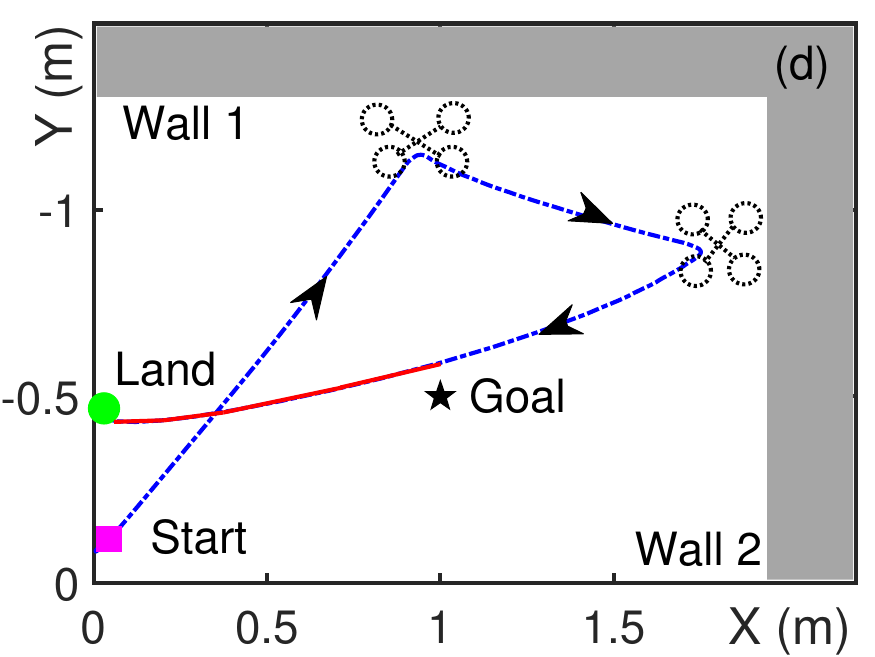}}
\caption{Results for a dual collision experiment with online mode switching demonstrate the collision model's accuracy in predicting peak contact forces (in inertial frame). The peak accelerations immediately following collisions are attributed to control efforts from the impact-agnostic low-level controller, which tracks new waypoints generated by the recovery controller.}
    \label{fig:validate_dual_collision}
\vspace{-0.2in}
\end{figure}

\subsection{Validation for Mode Selection}
In this section, we validate the results for the mode-selection case of Section \ref{sec:mode_selection} in a dual collision experiment. The first wooden wall is situated at $y = -1.1$m and the second acrylic wall at $x = 1.65$m. Forward simulations as described in Section \ref{sec:mode_selection} were carried out to obtain an optimal mode sequence with the optimal pre-collision velocities. To fly from a goal at (0,0) to a goal at (1,-0.5), the ideal collision velocities were found to be $\dot{x}^\dagger = -0.6$m/s and $\dot{y}^\dagger = 0.5$m/s in the flexible mode $(\Gamma = 2)$ which would result in a pre-collision velocity of $\dot{x}^\dagger = -1.6$m/s and $\dot{y}^\dagger = 0.25$m/s for the second collision. In order to reach the goal by using the recovery controller, the ideal mode for the second collision was found to be rigid $(\Gamma = 1)$. Accordingly, DART first approaches a wooden wall with its arms unlocked, in flexible configuration, and then upon collision, it engages its lock mechanism for a rigid configuration. It then approaches to collide with the acrylic wall in the rigid mode to validate the example case of Fig. \ref{fig:mode_selection} and as shown in Fig. \ref{fig:first_pic}. The peak impact forces which were predicted and the actual peak forces are plotted in Fig. \ref{fig:validate_dual_collision}. Similar to case in Section \ref{sec:pre-collision_vel}, the peak impacts and the trends are precisely predicted by our collision model for both $\lambda_{x^\dagger}$ and $\lambda_{y^\dagger}$. The contact force $\lambda_{z^\dagger}$ again shows some discrepancy, however keeping consistent when averaged over the entire contact period. The results for a dual collision significantly depend on the frictional contact force $\lambda_{y^\dagger}$ which varies in 2 out of 5 trials for the wooden wall due to the rough texture of the wall. However, collisions with acrylic wall are very consistent, indicating a need for future studies on the dependence of $\mu$ and $\nu$ on the materials involved in the collisions. 




\section{Conclusion} \label{sec:conclusion}
In this article, we developed DART, a novel collision-resilient drone with two stiffness modes that can be actively engaged in-flight using a servo mechanism. We thoroughly characterized DART's post-collision behavior for both settings and show that the two modes differ by a factor of 7 times in the body stiffness values. We also presented a novel 3D collision model based on LCS theory, which accurately predicted post-collision states for a variety of pre-collision conditions. Finally, we utilized this LCS-based collision model to demonstrate successful mode selection and pre-collision velocity selection for collision-inclusive trajectories. 

Future work will focus on developing planning algorithms by exploiting the proposed collision model to enable collision-inclusive maneuvers and design of low-level controllers to account for the impact forces.


\bibliography{bibliography.bib}
\bibliographystyle{ieeetr}

\end{document}